\pdfoutput=1

\documentclass[11pt]{article}

\usepackage[]{acl}

\usepackage{times}
\usepackage{latexsym}

\usepackage[T1]{fontenc}

\usepackage[utf8]{inputenc}

\usepackage{microtype}

\usepackage{graphicx}
\usepackage{url}
\usepackage{amsmath}
\usepackage{breqn}
\usepackage{makecell}
\usepackage{tabularx}

\usepackage{bm}
\newcommand{\specialcell}[2][c]{%
\begin{tabular}[#1]{@{}c@{}}#2\end{tabular}}

\title{Surfer100: Generating Surveys From Web Resources on Wikipedia-style}

\author{%
  Irene Li, Alexander Fabbri, Rina Kawamura, Yixin Liu$^*$, Xiangru Tang$^*$, \\
  \textbf{Jaesung Tae$^*$, Chang Shen$^*$, Sally Ma$^*$, Tomoe Mizutani$^*$, Dragomir Radev} \\
  Department of Computer Science\\
  Yale University \\
  $*$: authors mainly contributed to annotation.
  }

\begin{document}
\maketitle
\begin{abstract}
Fast-developing fields such as Artificial Intelligence (AI) often outpace the efforts of encyclopedic sources such as Wikipedia, which either do not completely cover recently-introduced topics or lack such content entirely. As a result, methods for automatically producing content are valuable tools to address this information overload. We show that recent advances in pretrained language modeling can be combined for a two-stage extractive and abstractive approach for Wikipedia lead paragraph generation. We extend this approach to generate longer Wikipedia-style summaries with sections and examine how such methods struggle in this application through detailed studies with 100 reference human-collected surveys. This is the first study on utilizing web resources for long Wikipedia-style summaries to the best of our knowledge.

\end{abstract}

\section{Introduction}

Novel concepts are being introduced and evolving at a rate that makes creating high-quality, up-to-date Wikipedia pages for such topics challenging. A pipeline for automatically creating such Wikipedia pages is thus desirable.  While there has been some work on generating full Wikipedia pages, these efforts are either domain-specific \cite{sauper2009automatically}, making strong assumptions about the topics being summarized \cite{banerjee2016wikiwrite}, or are purely extractive \cite{jha2015surveyor}. In a related line of work, query-based summarization has been applied to specific sections of Wikipedia pages \newcite{deutsch2019summary, zhu2019transforming}, which can be viewed as a more self-contained version of Wikipedia page generation.
Recent Wikipedia page generation work has focused on generating the initial leading paragraph of a Wikipedia page \cite{liu2018generating,liu2019hierarchical,perez2019generating}. These papers consist of a two-step framework by which an extractive method selects relevant content for a specific topic, and an abstractive method generates the final summary of the topic.
\par
In this paper, we first examine how recently-introduced pretrained language models \cite{devlin2019bert, liu2019roberta, lewis2019bart} improve upon both the extractive and abstractive steps of previous models for the task of lead paragraph generation. We further focus on analyzing the extension of such methods to full Wikipedia page generation on scientific topics related to AI and Natural Language Processing (NLP). We manually create summaries of 100 AI and NLP topics divided along sections, as on Wikipedia pages. We perform ablation studies on content selection and generation methods over selected topics, finding that current content selection methods are not precise and fail to differentiate content well among queries for subtopics of the main topic. 
\par
Our contributions are: 1) We demonstrate how recent advances in pretrained language models improve upon Wikipedia lead paragraph generation. 2) We then extend such a method to generate full Wikipedia-style pages of scientific topics; 3) For a testing purpose, we manually collected Surfer100, 100 SURveys  From wEb Resources on scientific topics, filling the gap on human-written surveys using web resources in scientific topics. We provide a better understanding of current methods and their faults on a real-world application. 

\section{Wikipedia Lead Paragraph Generation}
\vspace{-2mm}
In this section, we show how combining recent methods for a two-staged approach of content selection and generation give improved results on the WikiSum dataset \cite{liu2018generating} as well as a newly curated set of Wikipedia articles. 

\subsection{Data}

We make use of the \textbf{WikiSum} dataset \cite{liu2018generating}, a collection of over 1.5 million Wikipedia pages and their references. Applying pretraining techniques such as BERT \cite{devlin2019bert}, RoBERTa \cite{liu2019roberta}, and BART \cite{lewis2019bart}; however, these models make use of Wikipedia during pretraining. To address this problem, we mirror the process of \newcite{liu2018generating} to collect an unbiased dataset of newly added Wikipedia pages\footnote{https://en.wikipedia.org/wiki/Special:NewPages} which did not appear in pretraining, (\textbf{NewPage WikiSum}).  
We collect 10,000 of the newest Wikipedia pages, scrape Wikipedia for their references and the top 10 Google Search results. We remove non-English results and any articles for which we could not scrape a single reference. Due to the sparsity of search results on specific topics, we were left with about 1,000 articles we used as a test set. 

\subsection{Step One: Content Selection}

We experiment with five approaches for our initial content-selection step. \textbf{TF-IDF}: a simple approach to extract relevant content is to use term frequency–inverse document frequency \cite{liu2018generating, fan2019eli5}. \textbf{LSTM-Rank}: \newcite{liu2019hierarchical} approach query-based content selection as a regression problem of predicting the ROUGE-2 recall of a given paragraph-topic pair. \textbf{WikiCite}: \newcite{deutsch2019summary} approach query-based summarization via an extractive classification approach with attention \cite{bahdanau2014neural} over the topic and context. 
\par
We apply two additional methods to the task of content selection. \textbf{Semantic Search}: \newcite{reimers2019sentence} fine-tune BERT and Roberta using siamese and triplet networks to produce fixed-length vectors which can be compared using cosine similarity to find semantically similar input. 
We embed the title of each Wikipedia page, and each candidate paragraph, using this method, and choose the paragraphs with the most similar vectors to the title as selected content. 
\textbf{RoBERTa-Rank}: we train RoBERTa similar to the approach of \cite{liu2019hierarchical}, treating the title and paragraph to be ranked as sentence pairs and use predicted relevance scores as a ranking function for determining the most relevant paragraphs.
We show the results in Table \ref{tab:content_selection}. WikiCite performs well despite not including extensive pretraining and without fine-tuning on the WikiSum data, perhaps because the model is trained for the task of fine-grained selection (for section titles within a given page). RoBERTa-Rank is the highest-scoring content selector except for the 5-paragraph case, so then we choose this as the content selection method for abstractive summarization input on WikiSum data. 

\begin{table}[t!]
\centering
\small
\resizebox{\columnwidth}{!}{\begin{tabular}{l c c c c} \Xhline{2\arrayrulewidth}
             \specialcell{Methods} & \textbf{L=5} & \textbf{L=10} & \textbf{L=20} & \textbf{L=40} \\ \Xhline{\arrayrulewidth}
             TF-IDF & 24.86 & 32.43 & 40.87 &  49.49  \\  
              LSTM-Rank & 39.38 &  46.74 &  53.84 & 60.42  \\  
              WikiCite & \textbf{65.27} & 69.77 &  73.54 & 76.51  \\  
              Semantic Search & 34.87 & 48.60 & 61.87 & 74.54  \\ 
              RoBERTa-Rank & 64.12 & \textbf{72.49} & \textbf{79.17} & \textbf{84.28}  \\ \Xhline{2\arrayrulewidth}
\end{tabular}}
\caption{ROUGE-L-Recall scores for WikiSum content selection, varying the number of paragraphs returned.}
\label{tab:content_selection}
\vspace{-3mm}
\end{table}

\subsection{Step Two: Abstractive Summarization}
We use the RoBERTa-Rank content selection component to select paragraphs up to 1,024 total tokens as input to our abstractive summarization step. As the abstractive model in our two-step approach, we experiment with \textbf{BART} \cite{lewis2019bart}, which has achieved state-of-the-art performance in both natural language understanding and generation tasks. We compare BART fine-tuned on the WikiSum data with the previous state of the art \textbf{HierSumm} model from \citet{liu2019hierarchical}.
\par
We show improved results on generating the introduction paragraph on WikiSum and on our NewPage WikiSum data in Table \ref{tab:text_generation}. We use the same RoBERTa-Rank for both models on NewPage WikiSum. BART generation still outperforms HierSumm. We note that the large difference in scores between that of the WikiSum data and on our collected subset is likely due to the widespread nature of topics in WikiSum; WikiSum includes many well-established topics for which finding reference documents is simple, while the newly introduced topics may not contain enough reference information for higher-quality generation. So far, we have shown that applying RoBERTa-Rank and BART as a two-step pipeline gives promising results in generating lead Wikipedia sections. 

\begin{table}[t]
\centering
\small
\begin{tabular}{c c c } \Xhline{2\arrayrulewidth}
             \specialcell{Dataset} & \textbf{Hiersumm} & \textbf{BART} \\ \Xhline{\arrayrulewidth}
            WikiSum &  41.53/26.52/35.76 & \textbf{46.61}/\textbf{26.82}/\textbf{43.25} \\  
           NewPage  & 31.64/15.06/27.13 & \textbf{39.29}/\textbf{18.56}/\textbf{36.03} \\   \Xhline{2\arrayrulewidth}
\end{tabular}
\caption{ROUGE scores for intro paragraph generation on WikiSum and NewPage WikiSum.}
\label{tab:text_generation}
\vspace{-3mm}
\end{table}

\section{Application of Pipeline to Full Wikipedia Generation}
We follow \newcite{banerjee2016wikiwrite} in extending a two-step pipeline to full Wikipedia-style summaries (section by section content selection and summarization) to study the applicability of recent methods in this real-world setting. 
\subsection{Data}
Testing our models on full Wikipedia-page data would again face the problem of pretraining bias, and large-scale collection of full-size Wikipedia pages for novel topics is not infeasible. Furthermore, we focus on generating Wikipedia pages for AI-related topics. 
We picked a mixture of NLP and broader AI-related topics to include topics with existing Wikipedia pages as well as those without pages or stub articles, with 100 topics in total. 

\par
We define a template for the surveys consisting of five sections: \textbf{Introduction}, \textbf{History}, \textbf{Key Ideas}, \textbf{Variations} (similar topics or topics with similar goals) and \textbf{Applications}. We arrived at these section titles by an examination of sample Wikipedia pages in NLP. First, we searched Google for the given topic, retrieving all HTML page links for the first two search result pages. We then have the annotator read each page, extract relevant content into the corresponding section, and paraphrase and summarize the relevant content for each section to between 50 and 150 words per section. We split the job to eight annotators, and each survey requires 45 to 60 minutes. Given that the data collection is time-consuming, we focus on a testing purpose rather than training. We make all data public. \footnote{https://github.com/IreneZihuiLi/Surfer100} 
\subsection{Content Selection}
\label{sec:cs}
We first tested the quality of the content selection methods for generic retrieval of content relevant to a topic on our data. We choose the Semantic Search, WikiCite, and RoBERTa-Rank methods from Table \ref{tab:content_selection} for analysis. 
For Semantic Search, we experiment with three types of sentence embeddings, the original sentence-transformer BERT embeddings (\textbf{SS-BERT}), embeddings fine-tuned with SciBERT (\textbf{SS-SciBERT}), and a version fine-tuned to differentiate whether two paragraphs belong to the same Wikipedia section (\textbf{SS-Wiki}).  
Surprisingly, we found such content was often returned during retrieval despite the poor grammaticality and relevance.  We hypothesize that the tendency to return short sentences, often with odd punctuation may relate to the extension of these methods to paragraph levels while inherently being developed for sentence-level tasks.  
\par
Notably, the WikiCite method performs much better than semantic search and close to RoBERTa. We believe this is because the method is trained for content selection based on a topic and not simply trained for returning content with high recall. A potential problem with current methods in this two-step approach is that content selection is trained and evaluated with recall in mind, to capture as large a range of the topic, which produces models without the precision necessary in a real-world application. This aligns with previous work in extractive summarization suggesting that optimizing for recall gives suboptimal results \cite{zopf-etal-2018-scores}. 

\begin{table}[t]
\centering
\begin{tabular}{lccc}
\Xhline{2\arrayrulewidth}
    \specialcell{Methods}      & \textbf{k=10}   & \textbf{k=20}   & \textbf{k=50}   \\ \Xhline{\arrayrulewidth} 
SS-BERT & 0.5360 & 0.5370 & 0.4242 \\
SS-Wiki     & 0.5050 & 0.5125 & 0.4110 \\
SS-SciBERT   & 0.5780 & 0.5555 & 0.4232 \\
WikiCite        & \textbf{0.7460} & 0.6605 & 0.4722 \\
RoBERTa-Rank   & 0.7240 & \textbf{0.6925} & \textbf{0.5024} \\
    \Xhline{2\arrayrulewidth}
\end{tabular}
\caption{Evaluation on Content Selection: comparison of AvgP@k scores.}
\label{tab:precision_retrieval}
\end{table}

\par 
\textbf{Section-Specific Content Selection:}
We investigated the ability of our content selection models to retrieve content specific for each chosen section, for example, querying \lq \lq History of BERT\rq \rq rather than \lq \lq BERT.\rq \rq  We observed large overlaps between the returned results, between 5 and 9 paragraph overlap between the top 10 results for each section. Among all methods, Wikicite has the least overlap.
As an alternative method to select distinct content for each section, we investigate clustering methods, using out-of-the-box Agglomerative \cite{mullner2011modern} clustering provided by scikit-learn \footnote{https://scikit-learn.org/stable/index.html}. We cluster the embeddings obtained before the final output layer from the WikiCite and RoBERTa methods, and the Search-Wiki embeddings. We annotated the coherence of each cluster. Clusters obtained using embeddings from  RoBERTa, Search-Wiki and WikiCite had a corresponding average coherence of 3.07, 3.40, and 3.52 on a 1-5 scale, signaling slightly above-average coherence for each clustering. Again, the poor performance of RoBERTa in clustering may be due to the more general topic training method. As suggested by \newcite{deutsch2019summary}, the WikiCite method may dilute topic information in the final layer despite topic attention in previous layers and thus benefit from using embeddings before the final layer as clustering.

\subsection{Abstractive Summarization} 
\par
\textbf{Generation Model Choice:} To perform study on the choice of generation model, we took the best performing WikiCite and RoBERTa-Rank content selection methods for the introduction paragraph as input to BART. We also compared with classic baselines: FirstK (K=3,5) and TextRank \cite{mihalcea-tarau-2004-textrank} and MMR \cite{goldstein-carbonell-1998-summarization}. We show the ROUGE \cite{lin-2004-rouge} score on 100 topics in Table \ref{tab:rouge}. One can notice that, among the baselines, WikiCite has better performance among RoBERTa-Rank, marked with underlines. However, in the baselines, none of the summarization methods are robust. The best performance can be found when applying our pipeline (RoBERTa-Rank with BART) and this method surpasses other selected baselines in all cases.

\begin{table}[t!]
\centering
\small
\begin{tabular}{lccc}
\hline
    \specialcell{Method}   & \textbf{R-1}   & \textbf{R-2}   & \textbf{R-L}  \\ \Xhline{\arrayrulewidth}
WikiCite + First3    & 31.37 & 9.74  & \underline{20.51} \\
WikiCite + First5    & 32.53 & \underline{9.96}  & 20.30 \\
WikiCite + TextRank & \underline{32.57}& 9.79  & 19.03 \\
WikiCite + MMR & 29.79 & 6.66 & 16.82 \\
RoBERTa-Rank + First3     & 29.68 & 7.87  & 18.93 \\
RoBERTa-Rank + First5     & 30.64 & 7.95  & 18.71 \\
RoBERTa-Rank + TextRank  & 29.37 & 7.25  & 16.81 \\
RoBERTa-Rank + MMR & 28.78 & 4.71 & 15.33 \\
\Xhline{\arrayrulewidth}
WikiCite + BART     & 29.00 & 6.86  & 18.57 \\
RoBERTa-Rank + BART      & \textbf{32.23} & \textbf{10.12} & \textbf{21.78} \\
    \Xhline{2\arrayrulewidth}
\end{tabular}
\caption{Summarization performance: ROUGE scores.}
\label{tab:rouge}
\vspace{-3mm}
\end{table}

We further conduct human evaluation on WikiCite + BART and RoBERTa-Rank + BART. We randomly select 20 concepts and ask two human judges to give scores (range 1-5) on the following four perspectives: readability, relevancy, redundancy and  hallucination. For readability and relevancy, higher score is better; but for redundancy and  hallucination, higher score is not preferred, as we want the survey to be less redundant and hallucinate on the content. 
Results are shown in Table \ref{tab:human}. As seen in the Table, RoBERTa+BART performs better in the most cases, which is consistent with the ROUGE evaluation. Both models have a high hallucination score, we hypothesize that the content selection step keeps too many information that should not be included in the leading paragraph, for example, model technical details. 


\textbf{Generation of Full Summaries:}
We take the clustering output for the three embedding methods in the previous section (\textbf{Cluster Search-Wiki},\textbf{Cluster WikiCite}, and \textbf{Cluster RoBERTa}) as well as the Search-Wiki retrieval output(\textbf{Retrieval Search-Wiki}) as input to our generation component to create full sectioned summaries.
We did not conduct similar evaluation as we think the trend would be similar to the evaluation for the leading paragraph. Instead, we show a case study in Table \ref{tab:example}, and the full version can be found in Appendix. 
We could see there are descriptions about this topic: \lq \lq Text summarization is an interesting machine learning field \rq\rq, \lq \lq Automatic summarization aims to ... \rq\rq.   
We find certain stylistic features present in the surveys do not match Wikipedia pages. For example, some content is stated in the first person: \lq \lq In our new paper, we...\rq\rq.  This is an artifact of the generation model and the content extracted and can likely be remedied by fine-tuning BART in a different setting.

\begin{table}[t!]
\centering
\small
\begin{tabularx}{\linewidth}{|X|}
\hline
\textbf{Introduction}:
Text summarization is an interesting machine learning field that is increasingly gaining traction. As research in this area continues , we can expect to see breakthroughs that will assist in fluently and accurately shortening long text documents. In this article, we look at how machine learning can be used to help shorten text. \\
\hline

\textbf{Key Ideas}:
Automatic summarization aims to produce a shorter version of an input text, preserving only the essential information. There are two main types of summarization : extractive summarization selects important sentences from the input and abstractive summarizing generates content without explicitly re-using whole sentences. In our new paper , we constructed two novel , large-scale summarization datasets from scientific journal articles.  \\

\hline
\textbf{Applications}:
Summarization can be a crucial component in the tele-health supply chain when it comes to analyzing medical cases. The Spreading Activation approach does not allow to improve our results. Tables 8 and 9 show the high recall obtained with these methods, which may be a very interesting feature in some cases. \\
\hline
\end{tabularx}
\caption{Sample survey (part) of the topic \texttt{Text Summarization} created using our pipeline. }
\label{tab:example}
\end{table}


\begin{table}[t]
\centering
\small
\begin{tabular}{lcc}
\hline
  \specialcell{Evaluation} & WikiCite+BART    & RoBERTa+BART \\ \Xhline{\arrayrulewidth} 
Readability  & 3.70     & \textbf{4.15}       \\
Relevancy      & 3.28     & \textbf{3.58}     \\
Redundancy*     & \textbf{1.40}    & 1.43     \\
Hallucination*   & 2.65  & \textbf{2.55}      \\
\Xhline{2\arrayrulewidth}
\end{tabular}
\caption{Human evaluation on the leading paragraph generation, average scores on 20 random selected topics. * means lower score is better. RoBERTa is a short form of RoBERTa-Rank.}
\label{tab:human}
\end{table}

\section{Conclusion}
In this paper we show improvements in individual components of Wikipedia summarization through an application of recently-introduced embedding and summarization techniques, but largely focus on the failures of these methods when extended in a real-world scenario of full-page Wikipedia-styled summarization. We believe that a focus on high-precision and fine-grained query-based summarization in future work will help make this pipeline viable.

\bibliography{anthology,custom}
\bibliographystyle{acl_natbib}
\clearpage
\appendix


\section{Dataset and Training Details}
For training RoBERTa-Rank, we sampled 1,209,387 training and 10,000 validation paragraph from the original WikiSum dataset.
\par
For training the WikiSum component, we took a subset of the original WikiSum dataset consisting of 280,000 training instances and 10,000 validation instances. We removed paragraphs which were clones of the target summary through a threshold of .5 ROUGE-2 score. We then sort the instances according to the sum of the ROUGE scores of individual paragraphs and take the paragraphs with highest scores for training and validation. This was done to filter out examples with poorly collected source documents and promote a stronger connection between the source documents and target summary. The number of training examples was chosen to be close to the number found in the CNN-DailyMail dataset.  
\par
We are releasing all data used in this paper, include the NewPages WikiSum and the data used in our ablation studies. 
\par
For training BART on the above WikiSum data, we train with a polynomial decay learning rate scheduler with learning rate $3\mathrm{e}{-5}$, using the Adam optimizer \citep{kingma2014adam}. We train with 500 warmup steps and 20,000 total steps, ending with a validation loss of 3.492. The max-tokens per batch is 1024 and update frequency for gradient accumulation is 8. Additional details can be obtained from the checkpointed model released with the paper. The model is the same as the BART large model released by Facebook, without any additional parameters, for a total of 405,766,144 parameters. This model was trained on 8 16 GB V100 GPUs for about 10 hours. 
\par
For training RoBERTa-Rank, we train with a polynomial decay learning rate scheduler with learning rate $2\mathrm{e}{-5}$, using the Adam optimizer \citep{kingma2014adam}. We train with 6000 warmup steps and 10,0000 total steps, where by the end of training the validation loss is practically 0. The model has 356,461,658 parameters, building off of RoBERTa large. Additional details can be obtained from the checkpointed model released with the paper. This model was also trained on 8 16 GB V100 GPUs for about a day.

\section{Topics for Analysis}
Below we list the topics used in our analysis. 
\begin{table}[h!] 
\small
    \centering
\begin{tabular}{c} \hline \hline
              \textbf{Topics}   \\ \hline
adagrad (optimizer)		\\ 
adam (optimizer)	\\
attention mechanism (deep learning)		\\
BERT		\\
convolutional neural networks		\\
image captioning (deep learning)	\\
knowledge graphs		\\
recursive neural networks		\\
rmsprop (optimizer)		\\
sentiment analysis		\\
\hline \hline
\end{tabular}
\caption{A list of the topics used for ablation studies (Section \ref{sec:cs}, Selection-Specific Content Selection).}
\label{tab:topics}
\end{table}

\begin{table}[h!] 
\small
    \centering
\begin{tabular}{c} \hline \hline
              \textbf{Topics}   \\ 
              \hline
maximum marginal relevance \\
perceptron \\
sentiment analysis \\
language modeling \\
autoencoders \\
gaussian mixture model \\
ensemble learning \\
long short-term memory network \\
gradient boosting \\
meta learning \\
residual neural network \\
multilingual BERT \\
hidden markov models \\
clustering \\
decision trees \\
relation extraction \\
BERT \\
knowledge graphs \\
generative adversarial network \\
rmsprop optimizer \\
\hline \hline
\end{tabular}
\caption{Randomly selected 20 topics used for human evaluation (Table \ref{tab:human}).}
\label{tab:topics_test}
\end{table}

\section{Example Summaries}
Below we show examples of the topic summaries created by our pipeline, using RoBERTa embeddings for clustering, a CNN-DailyMail BART checkpoint for summarization and manually assigning associated section labels.
\begin{table}[h]
\centering
\small
\begin{tabularx}{\columnwidth}{|X|}
\hline
\textbf{Introduction} \\ \hline
Text summarization is an interesting machine learning field that is increasingly gaining traction. As research in this area continues , we can expect to see breakthroughs that will assist in fluently and accurately shortening long text documents. In this article, we look at how machine learning can be used to help shorten text. \\
\hline
\textbf{History} \\ \hline
Summarization has been and continues to be a hot research topic in the data science arena. While text summarization algorithms have existed for a while , major advances in natural language processing and deep learning have been made in recent years. Google has reportedly worked on projects that attempt to understand novels. Summarization can help consumers quickly understand what a book is about.  \\
\hline

\hline
\textbf{Key Ideas} \\ \hline
Automatic summarization aims to produce a shorter version of an input text, preserving only the essential information. There are two main types of summarization : extractive summarization selects important sentences from the input and abstractive summarizing generates content without explicitly re-using whole sentences. In our new paper , we constructed two novel , large-scale summarization datasets from scientific journal articles.  \\

\hline
\textbf{Variations} \\ \hline
Multi-document summarization can be a powerful tool to quickly analyze dozens of search results. MeaningCloud 's Summarization API locates the most relevant phrases in a document and builds a synopsis with them. More specific summarization systems could be developed to analyze legal documents. \\
\hline

\hline
\textbf{Applications} \\ \hline
Summarization can be a crucial component in the tele-health supply chain when it comes to analyzing medical cases. The Spreading Activation approach does not allow to improve our results. Tables 8 and 9 show the high recall obtained with these methods, which may be a very interesting feature in some cases. \\
\hline
\end{tabularx}
\caption{Sample survey of the topic \texttt{Text Summarization} created using our automated pipeline, showing both the ability of our pipeline to capture important content as well as problems related to the style of presentation, such as references to input Tables.}
\label{tab:example1}
\end{table}

\begin{table}[t]
\centering
\small
\begin{tabularx}{\columnwidth}{|X|}
\hline
\textbf{Introduction} \\ \hline
Dropout is a technique where randomly selected neurons are ignored during training. This means that their contribution to the activation of downstream neurons is removed. Dropout alone does not have any way to prevent parameter values from becoming too large during this update phase. In the example below we add a new Dropout layer between the input ( or visible layer ) and the first hidden layer. The dropout rate is set to 20\%, meaning one in 5 inputs will be randomly excluded from each update cycle.
\\
\hline
\textbf{History} \\ \hline
Classical generalization theory suggests that to close the gap between train and test performance , we should aim for a simple model. Christopher Bishop formalized this idea when he proved that training with input noise is equivalent to Tikhonov regularization. In 2014, Srivastava et al. developed a clever idea for how to apply Bishop 's idea to the internal layers of the network. They proposed to inject noise into each layer of the Network before calculating the subsequent layer.
 \\
\hline

\hline
\textbf{Key Ideas} \\ \hline
Additionally , as recommended in the original paper on Dropout , a constraint is imposed on the weights for each hidden layer. This is done by setting the kernel_constraint argument on the Dense class when constructing the layers. In the example below Dropout is applied between the two hidden layers and between the last hidden layer and the output layer.
  \\

\hline
\textbf{Variations} \\ \hline
With a Gaussian-Dropout , the expected value of the activation remains unchanged. Unlike the regular Dropout , no weight scaling is required during inferencing. Dropout is only used during the training of a model and is not used when evaluating the skill of the model. The main problem hindering dropout in NLP has been that it could not be applied to recurrent connections.
\\
\hline

\hline
\textbf{Applications} \\ \hline
During training time , dropout randomly sets node values to zero. During inference time, dropout does not kill node values, but all the weights in the layer were multiplied. This multiplier could be placed on the input values rather than the weights. TensorFlow has its own implementation of dropout which only does work during training time.
\\
\hline
\end{tabularx}
\caption{Sample survey of the topic of \texttt{Dropout}. Some stylistic problems such as references to examples described in the original document are present, although key concepts of the topic are addressed.}
\label{tab:example2}
\end{table}

\end{document}